# A Single Euler Number Feature for Multi-font Multi-size Kannada Numeral Recognition

B.V.Dhandra, R.G.Benne and Mallikarjun Hangarge
P.G .Department of Studies and Research in Computer Science,
Gulbarga University, Gulbarga
INDIA
dhandra_b_v@yahoo.co.in, rgbenne@yahoo.com

**ABSTRACT**

*In this paper a novel approach is proposed based on single Euler number feature which is free from thinning and size normalization for multi-font and multi-size Kannada numeral recognition system. A nearest neighbor classification is used for classification of Kannada numerals by considering the Euclidian distance. A total 1500 numeral images with different font sizes between (10..84) are tested for algorithm efficiency and the overall the classification accuracy is found to be* 99.00% *.The said method is thinning free, fast, and showed encouraging results on varying font styles and sizes of Kannada numerals.*

## 1. Introduction

Automatic numeral recognition system is an important component of character recognition system due to its vide applications field like reading postal zip code, number plate of vehicle, passport number, employee code, bank check, and form processing. The problem of printed multi-font multi-size numeral recognition is difficult task due to the variations of font style and font sizes of numerals.

The problem of numeral recognition has been attempted for last one decade and many methods have been proposed such as template matching, dynamic programming, hidden Markov modeling, neural network, expert system and combinations of all these techniques [1,7,8]. A various feature extraction methods for character recognition is presented by Ivind and Jain[4]. Extensive work has been carried out for recognition of character/numeral in foreign languages like English, Chinese, Japanese, and Arabic. In the Indian context some major works are reported on Devanagari, Tamil, Bengali, and Kannada numeral recognition [2,3]. Dinesh Acharya *et. al*[5] uses 10-segment string, water reservoir, horizontal and vertical stroke features for Kannada numeral recognition, U.Pal *et. al* [10] uses zoning, directional chain code for Kannada numerals recognition. Sanjeev Kunte and Sudhakar Samuel[11] has presented script independent handwritten Numeral recognition system with wavelet feature and neural network classifier. Dhandra *et. al*[6,12] have proposed a thinning free multi font multi size English numeral recognition based on directional density. From the literature it reveals that there are methods which are suffers from larger computation time mainly due feature extraction for large set and various preprocessing stages, i.e. size normalization, and skeletonning or thinning of an images. Also recognition system fails to meet the desired accuracy when it is exposed to the different font sizes and styles. Hence, it is necessary to develop a thinning free method which is independent of size, style, and has high recognition rate. This has motivated us to design a simple, efficient, and robust algorithm for Kannada numerals recognition system.

In this paper, a simple, fast, thinning free novel method is proposed for multi-size multi-font Kannada numeral recognition system without size normalization. A single Euler number feature is used for Kannada numeral recognition. The Euler numbers obtained from an image, vertical zoned image and horizontal zoned image are considered for recognition.

The paper is organized as follows: Section 2 contains the Kannada numerals and preprocessing. Feature Extraction Method is described in Section 3. The proposed algorithm is presented in Section 4. The Classification methods are subject matter of Section 5. The experimental details and results obtained are presented in Section 6 and section 7 contains the conclusion.

## 2. Kannada numerals and pre-processing

Kannada language is one among the four major south Indian Languages spoken by about 50 million people. The Kannada alphabet consists of 16 vowels and 36 consonants and 10 numerals. The sample numerals are given in Table 1.

**Table 1: Kannada Numerals**

| Eng. | 0 | 1 | 2 | 3 | 4 | 5 | 6 | 7 | 8 | 9 |
|---|---|---|---|---|---|---|---|---|---|---|
| Kan. | ೦ | ೧ | ೨ | ೩ | ೪ | ೫ | ೬ | ೭ | ೮ | ೯ |

**Table 2: A sample data set of printed Kannada numerals**

| Sample Numerals | Font type |
|---|---|
| ೦೧೨೩೪೫೬೭೮೯ | BRH-Kannada |
| ೦೧೨೩೪೫೬೭೮೯ | BRH-Amerikannada |
| ೦೧೨೩೪೫೬೭೮೯ | BRH-Kailasm |
| ೦೧೨೩೪೫೬೭೮೯ | BRH-Vijay |
| ೦೧೨೩೪೫೬೭೮೯ | BRH-Kasturi |
| ೦೧೨೩೪೫೬೭೮೯ | BRH-Bangaluru |
| ೦೧೨೩೪೫೬೭೮೯ | BRH-Sirigannada |
| ೦೧೨೩೪೫೬೭೮೯ | BRH-Kannada Extra |
| ೦೧೨೩೪೫೬೭೮೯ | KGP_kbd |
| ೦೧೨೩೪೫೬೭೮೯ | Nudi Akshara-01 |
| ೦೧೨೩೪೫೬೭೮೯ | Nudi Akshara-02 |
| ೦೧೨೩೪೫೬೭೮೯ | Nudi Akshara-03 |
| ೦೧೨೩೪೫೬೭೮೯ | Nudi Akshara-04 |
| ೦೧೨೩೪೫೬೭೮೯ | Nudi Akshara-05 |
| ೦೧೨೩೪೫೬೭೮೯ | Nudi Akshara-06 |
| ೦೧೨೩೪೫೬೭೮೯ | Nudi Akshara-07 |
| ೦೧೨೩೪೫೬೭೮೯ | Nudi Akshara-08 |
| ೦೧೨೩೪೫೬೭೮೯ | Nudi Akshara-09 |
| ೦೧೨೩೪೫೬೭೮೯ | Nudi B-Akshara |
| ೦೧೨೩೪೫೬೭೮೯ | Nudi Akshara |

Data is collected from Baraha and Nudi software; for different font styles and font sizes between (10,…,84). The printed page containing multiple lines of isolated Kannada numeral is scanned through a flat bed HP scanner at 300 DPI and binarized using global threshold stored in bmp file format. The scanning artefacts are removed by using morphological opening operation. A sample data set of Kannada numerals and corresponding font styles are presented in Table 2.

## 3. Feature Extraction Method

Feature extraction and selection is an important component of any recognition system. Selection of feature is probably the single most important factor in achieving high recognition performance. Feature extraction is the identification of appropriate measures to characterize the component images distinctly. The proposed feature extraction technique is based on Euler number. The Euler number is obtained from an image, horizontal zoned image, and vertical zoned image as shown in figure 2.

### 3.1 Euler number:

Euler number of an image is defined as the number of objects in the region minus the number of holes in the objects [9]. Numeral '0','4','6',and '9' will have the Euler number 0 and '1','2','3','5',and'7'will have Euler number 1.

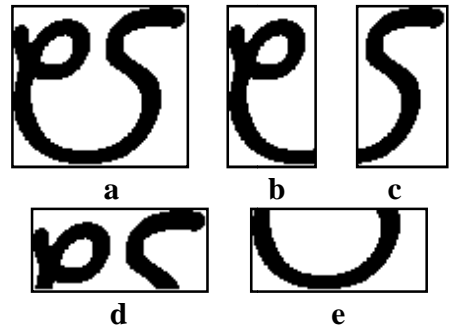

**Figure 2 (a) Kannada Numeral 8**
**Figure 2(b)-2(c) vertically divided into two zones**
**Figure 2(d)-2(e) horizontally divided into two zones**

## 4. Algorithm

**Input**   : Isolated Binary Kannada Numeral.

**Output**: Recognition of the Numeral.

**Method:** Euler number feature and k-NN classifier.

1. Preprocess the input image to eliminate the scanning artifacts using morphological opening operation and invert the image.

2. Fit the minimum rectangle bounding box for an input image and crop the digit.

3. Find the Euler number of an image, vertical zoned images and horizontal zoned images.

4. Estimate the minimum distance between feature vector and vector stored in the library by using Euclidian distances.

5. Classify the input image into appropriate class label using minimum distance K-nearest neighbor classifier.

6. Stop.

## 5 Classification

*K-Nearest-Neighbor (KNN) classifier*: Nearest neighbor classifier is an effective technique for classification problems in which the pattern classes exhibit reasonably limited degree of variability. The k-NN classifier is based on the assumption that the classification of an instance is most similar to the classification of other instances that are nearby in the vector space. It works by calculating the distances between one input patterns with the training patterns. A k-Nearest-Neighbor classifier takes into account only the k nearest prototypes to the input pattern.  Usually, the decision is determined by the majority of class values of the k neighbors.  It however, suffers from the twin problems of high computational cost and memory. K-nearest neighbor is more general than the nearest neighbor. In other words, nearest neighbor is a special case of k-nearest-neighbor, for k=1. The algorithm is executed for k=1, k=3, k=5 and k=7 the results are compared to find out the optimum value of k. From Table 3 it is clear that k=1 is optimal value.

In the k-Nearest neighbor classification, the distance between features of the test sample and the feature of every training sample are computed. The class of majority among the k-nearest training samples is based on the   Euclidian measures.

**Table 3:**
**The average recognition rate using different values of k with KNN classifier**

| K-NN classifiers | Recognition accuracy in % |
|---|---|
| K=1 | **99.00** |
| K=3 | 98.40 |
| K=5 | 97.10 |
| K=7 | 95.90 |

**Table 4:**
**The recognition results for Kannada numerals**

| numerals | Train images | Test images | % of recognition |
|---|---|---|---|
| ೦ | 50 | 25 | 100 |
| ೧ | 50 | 25 | 94 |
| ೨ | 50 | 25 | 100 |
| ೩ | 50 | 25 | 100 |
| ೪ | 50 | 25 | 100 |
| ೫ | 50 | 25 | 100 |
| ೬ | 50 | 25 | 100 |
| ೭ | 50 | 25 | 100 |
| ೮ | 50 | 25 | 96 |
| ೯ | 50 | 25 | 100 |
| Average recognition | | | 99.00 |

## 6 Results and discussion

The experiment is carried out on the data set of 750 images for each font style. For every font style, a random sample of 500 numerals for training and 250 numerals for testing are considered and the results are obtained. The table 4 shows the recognition rate of 99.00 % for recognition of Kannada numerals using K-NN classifier; The table 3 shows the average recognition results for k=1, 3, 5, and 7. The proposed method yields reasonable recognition accuracy with smaller feature set and basic classifier. However, the recognition rate can be improved by adding new potential features.

Table 5 presents the comparison of the proposed method with other methods in the literature dealing with Kannada numerals.

**Table 5:**
**Comparison of results with other methods**

| Method | Features, Classifier and script | Data set | % of Acc. |
|---|---|---|---|
| 6 | Structural features(13)- Directional density, Water reservoir, Max. Profile, fill hole density. K-NN classifier. Kannada Numeral | 1150 | 100 |
| 12 | Structural feature(5)- Directional density, Euler number, two stage classification with NN classifier, English Numeral | 1150 | 99.78 |
| Proposed | Topological feature (1)- Euler feature, K-NN classifier | 1500 | 99.00 |

The proposed method yields a comparable recognition rate of 99.00% with single feature taking at lesser computation cost even with the basic classifier.

## 7. Conclusions

In this paper only one topological feature is used for recognition of printed Kannada numerals. In any recognition process, the importance will be given to the feature extraction, dimensionality and suitable classification. The proposed algorithm attempts to address the above factors and performs better in terms of accuracy and time complexity. The Overall accuracy of 99.00% is achieved. The proposed method is free from thinning, size normalization, fast, and accurate. This work is carried out as an initial attempt, and the aim of the paper is to facilitate for robust Kannada OCR

### Acknowledgement

The authors thank the referees for their comments and suggestions. The authors wish to acknowledge Dr. P.Nagabhushan, Department of Computer Science, University of Mysore, Mysore, for suggesting the single novel feature idea for Kannada numeral recognition system.